\documentclass[a4paper,fleqn]{cas-dc}

\usepackage[numbers]{natbib}
\usepackage{array}
\usepackage{textcomp}
\usepackage{stfloats}
\usepackage{url}
\usepackage{verbatim}
\usepackage{graphicx}
\usepackage{hyperref}
\usepackage{algorithmic}
\usepackage{caption}
\usepackage{textcomp}
\usepackage{multirow}
\usepackage{subcaption}
\usepackage{verbatim}
\usepackage{multicol}
\usepackage{tabularx}
\usepackage{makecell}
\usepackage{wasysym}
\usepackage{booktabs}
\usepackage{rotating}
\usepackage{xcolor}
\def\tsc#1{\csdef{#1}{\textsc{\lowercase{#1}}\xspace}}
\tsc{WGM}
\tsc{QE}
\tsc{EP}
\tsc{PMS}
\tsc{BEC}
\tsc{DE}

\begin{document}
\let\WriteBookmarks\relax
\def\floatpagepagefraction{1}
\def\textpagefraction{.001}
\shorttitle{Data-driven Approach for Joint Entity and Relation Extractions}
\shortauthors{Mouiche and Saad}

\title [mode = title]{TIJERE: A Novel Threat Intelligence Joint Extraction Model Based on Analyst Expert Knowledge}                      



\author[1]{Inoussa Mouiche}[ 
                        orcid=0009-0008-8024-7631]
 \cormark[1]
\ead{mouiche@uwindsor.ca}

\credit{Conceptualization, Methodology, Data curation, Software, Validation,  Formal Analysis, Writing – Original draft preparation, Project administration}

\affiliation[1]{organization={School of Computer Science, University of Windsor},
                addressline={401 Sunset Ave}, 
                city={Windsor},
                postcode={N9B 3P4}, 
                state={ON},
                country={Canada}}


\author[1]{Sherif Saad}[%
   ]
\ead{shsaad@uwindsor.ca}

\credit{Conceptualization, Visualization, Investigation, Supervision, Writing – Reviewing and Editing}




\cortext[cor1]{Corresponding author}


\begin{abstract}
The extraction of entities and relationships from threat intelligence reports into structured formats, such as cybersecurity knowledge graphs, is essential for automated threat analysis, detection, and mitigation. However, existing joint extraction methods struggle with feature confusion, language ambiguity, noise propagation, and overlapping relations, resulting in low accuracy and poor model performance. This paper presents TIJERE, an innovative joint entity and relation extraction framework that formulates joint extraction as a multisequence labeling representation (MSLR) problem. Specifically, separate sequences are generated for each entity pair. Unlike prior tagging schemes, MSLR integrates expert domain features to enrich positional, contextual, and semantic representations of entities, thereby enhancing feature distinction and classification accuracy. Additionally, TIJERE reduces language ambiguity and enhances domain-specific generalization by leveraging SecureBERT+, a contextual language model fine-tuned on cybersecurity text. This improves both named entity recognition (NER) and relation extraction (RE). This paper also introduces DNRTI-JE, the first publicly available jointly labeled dataset for cybersecurity entity and RE, filling a crucial gap in cyber threat intelligence automation. Empirical evaluations on the curated DNRTI-JE dataset demonstrate that TIJERE achieves state-of-the-art performance, with F1-scores exceeding 0.93 for NER and 0.98 for RE, outperforming existing methods. Together, TIJERE and the standardized benchmarking DNRTI-JE dataset enable high-performance cybersecurity intelligence extraction, with transferable applications in healthcare, finance, and bioinformatics.
\end{abstract}


\begin{highlights}
\item TIJERE proposed for joint entity–relation extraction in cybersecurity.
\item Joint extraction framed as multisequence labeling for overlapping relations.
\item Expert-driven features used to separate entities and relations effectively.
\item SecureBERT$^+$ enhances domain-specific contextual representation.
\item DNRTI-JE is the first public dataset for joint NER and RE in CTI.
\end{highlights}

\begin{keywords}
Threat intelligence joint entity and relation extraction \sep
 Multisequence labeling representation \sep
 Expert domain features\sep
 Cyber threat intelligence\sep
 Cyber knowledge graphs \sep
 Pipeline extraction \sep
 Joint extraction
\end{keywords}

\maketitle

\section{Introduction}
Cyber threat intelligence (CTI) provides actionable insights that help organizations detect, prevent, and respond to cyberattacks. It involves analyzing unstructured threat reports to extract adversarial tactics, techniques, and procedures (TTPs), which are then represented in structured formats, such as cybersecurity knowledge graphs (CKGs). CKGs unify data from diverse sources to enhance threat landscape understanding and automate proactive defense mechanisms, particularly against advanced persistent threats (APTs)~\cite{OpenCyKG,HuikangZhang}. They enable security analysts to query threats, correlate indicators, predict attacks, and develop effective countermeasures. Constructing CKGs from CTI reports entails two core tasks: named entity recognition (NER) and relation extraction (RE). NER detects entities such as hacker groups, malware, and vulnerabilities, while RE predicts relationships between potential entity pairs. The extracted entity-relation pairs, known as relation triples, are represented as (E1, R12, E2), where E1 and E2 represent entity mentions, and R12 is the relation connecting them. These triples are stored in graph databases such as Neo4j to build the CKG. Two extraction paradigms exist: pipeline extraction (PE), which separates NER and RE into distinct models, and joint extraction (JE), which combines both tasks in a unified model~\cite{TiKG,CyberRel,CyberEntRel}.

Recently, the JE technique has become increasingly popular for its ability to model semantic interactions between NER and RE during training, thereby reducing the error propagation commonly encountered in JE settings~\cite{CyberRel,CyberEntRel,ZuoYali,XiaodiWang}. While JE improves performance in theory, existing JE methods in threat intelligence often fall short owing to several practical challenges:
\newline \textbf{Overlapping relations:} A sentence may contain multiple entity pairs that involve the same entities in different roles. JE models generally predict only one relation per pair, making it difficult to disambiguate overlapping relations~\cite{CyberRel,CyberEntRel,ZuoYali,ITIRel}.
\newline \textbf{Feature confusion:} When entity and relation features are not clearly separated, joint models struggle to learn both tasks effectively, particularly when sentence duplication is used to manage nested or overlapping structures, which introduces noise and increases classification errors~\cite{XiaodiWang,PipeVsJoint,ZexuanZhong}.
\newline \textbf{Ambiguous relation types:} Non-vocabulary expressions, such as "hasAttackTime" used to link a "hacker group" to an "attack time", are commonly employed by analysts to denote relationships between entities. While these terms are intuitive for humans, they pose semantic challenges for models that lack contextual understanding~\cite{XiaodiWang}.
\newline \textbf{Language context:} Generic language models are not trained in domain-specific cybersecurity jargon, resulting in poor entity detection and propagation of noise throughout the pipeline~\cite{TiKG,CTiKG}.

To address these challenges, \cite{ZuoYali} and \cite{CyberEntRel} introduced external relation-matching rules to enhance classification. However, their models effectively resemble pipeline architectures, as they treat NER and RE separately. Wang et al.~\cite{XiaodiWang} manually replaced ambiguous relation types with explicit terms at the embedding level; however, this approach is subjective and error-prone, particularly in large datasets. Similarly, Lv et al.\cite{CTI-TFN} proposed a task-specific model using separate BiLSTM-based Fourier networks for NER and RE. While enabling task-specific learning, it struggles with overlapping and complex relation types, which limits its performance and generalization. Overall, these studies highlight that RE remains the most difficult component in JE for cybersecurity.\\
This paper introduces TIJERE, a data-driven JE framework tailored for cybersecurity. TIJERE reformulates JE as a multisequence labeling problem, generating a multisequence labeling representation (MSLR) that addresses overlapping entities and relations. MSLR incorporates expert-driven features encoding positional and contextual semantics to enable effective interaction between NER and RE tasks while reducing feature confusion during training. This design enables the model to make more accurate predictions during inference. TIJERE also leverages SecureBERT$^+$\cite{SecureBERTPlus}, a cybersecurity-specialized variant of SecureBERT\cite{SecureBERT}, to better capture domain-specific language patterns and improve entity recognition. To support training and evaluation, we curate DNRTI-JE, the first publicly available dataset for joint entity-relation extraction in cybersecurity. It extends the original DNRTI dataset~\cite{DNRTI}, which was initially intended for NER. Following the methodology in~\cite{TiKG}, the authors predefined 15 relation types and a domain ontology schema to annotate DNRTI with relations, resulting in DNRTI-JE. This dataset includes 13 entity categories and 15 relation types across 6,592 sentences describing APT groups. DNRTI-JE is used to evaluate TIJERE against recent state-of-the-art approaches.

The key contributions of this paper are as follows:
\begin{itemize} 
\item We propose TIJERE, a novel architecture for jointly extracting entities and their relations from APT reports to construct CKGs. TIJERE formulates the task as a multisequence labeling problem using MSLR, where separate sequences are generated per entity pair to address overlapping relations. It integrates expert domain features, entity mask and entity type, to enrich semantic and positional context, improving feature distinction even in the presence of ambiguous relation types. Additionally, TIJERE leverages SecureBERT$^+$, a cybersecurity-specific language model, which enhances over general embeddings by reducing language ambiguity and better capturing domain-specific terminologies, thereby enhancing entity recognition and relation classification.
\item We introduce DNRTI-JE, a jointly labeled dataset for threat intelligence, which, to the best of our knowledge, is the first publicly available dataset designed for both NER and RE in cybersecurity. DNRTI-JE fills a critical gap in cybersecurity NLP research and is made available on our GitHub\footnote{https://github.com/imouiche/TIJERE} to support reproducibility and future research in this field.
\item We validate the effectiveness of TIJERE by conducting an ablation study and comparative evaluation against state-of-the-art approaches, all implemented from scratch using the DNRTI-JE dataset. Furthermore, we demonstrate the individual contributions of the engineered expert features on overall model performance.
\end{itemize}

To the best of our knowledge, this is the first study to propose a data-driven approach designed specifically to address JE challenges in the cybersecurity domain. The remainder of this paper is structured as follows: Section 2 reviews related studies and identifies key gaps in the literature. Section 3 details our proposed JE model, presents the evaluation results, and discusses its contributions against existing approaches. Finally, Section 4 summarizes the findings and outlines directions for future research.
\section{Related Works}
This section reviews peer-reviewed studies that have employed either PE or JE techniques for extracting entities and their relationships from unstructured cybersecurity reports for constructing CKGs.
\subsection{Pipeline Extraction Models}
In this approach, the NER and RE tasks are divided into two distinct modules.

Gasmi et al.~\cite{HoussemGasmi} employed a bidirectional long short-term memory (BiLSTM) model stacked with a conditional random field (CRF) for both entity recognition and relation classification tasks. Although the same model was utilized, the tasks were separated into different modules to achieve comparative results. Zhao et al.~\cite{JunZhao} employed an attention-based BiLSTM-CRF model to extract cyber entities, specifically indicators of compromise (IoCs), from security articles. Moreover, they developed a heterogeneous information network to identify predefined relationships between IoCs pairs. Jo et al.~\cite{Vulcan} developed an API framework called Vulcan, which automates the extraction of malware-related entities and their interconnections from APT reports. Vulcan combines a pretrained BERT model with BiLSTM and CRF, demonstrating significant performance improvements in basic operations such as threat evolution and profiling. Marchiori et al.~\cite{STIXnet} introduced the STIXnet pipeline to extract STIX objects and their relationships from APT reports. STIXnet integrates IoC Finder, rcATT, and regex for NER, along with neural networks (NNs) and a dependency parser for RE. Similarly, Li et al.~\cite{YongfeiLi} developed the entity extraction with multi-head attention and part-of-speech framework to perform four tasks: NER, coreference resolution, RE, and KG construction. Sarhan and Spruit~\cite{OpenCyKG} developed the open CTI knowledge graph framework, which contains two modules. The first module implements an attention-based neural open information extraction model to generate relation triples from CTI reports. The second module utilizes a BiGRU-CRF model for NER, which labels the extracted relation triples and populates the KG via a fusion technique. Piplai et al.~\cite{AritranPiplai} developed a pipeline model that combines regular expressions (regex) and CRF for entity recognition, along with RelExt-based NNs for relation classification. This framework automatically extracts entity-relation triples from after-action reports and populates the CKG using the unified cybersecurity ontology (UCO) 2.0. Recently, Mouiche and Saad~\cite{TiKG,CTiKG} introduced TiKG and CTiKG, novel PE frameworks for threat intelligence KGs. The aforementioned frameworks leverage state-of-the-art approaches, including SecureBERT embeddings, which capture security context to minimize classification errors and enhance the accuracy of the model.

The principal limitation of the above studies is their reliance on the pipeline approach, which suffers from error propagation owing to the separation of NER and RE into independent modules. Misclassification errors in entity recognition can cascade into RE, resulting in suboptimal performance and degraded KG quality~\cite{CyberEntRel,CyberRel,ZuoYali,XiaodiWang}. Contrastingly, this study implements a JE approach, integrating NER and RE within a unified model to enhance task interaction and minimize noise propagation, ultimately enhancing the overall performance of the model. Although a detailed comparison between JE and PE lies beyond the scope of this work, it should be acknowledged that the cybersecurity domain lacks comprehensive studies that validate this claim. Nevertheless, the development of joint techniques for the simultaneous extraction of entities and relations from unstructured data has garnered significant attention.
\subsection{Joint Extraction Models}
The joint technique integrates NER and RE into a unified model that facilitates their interaction to minimize error propagation and enhance classification accuracy. 
\begin{table*}[h]
\centering
\caption{Relevant studies on entities and relations extraction in cybersecurity using JE. JE: Joint extraction; Att: Attention; EDF: Expert domain features; MSLR: Mult-sequence labeling representation.}
\begin{tabular}{p{1.3cm}|c|p{1.3cm}|c|p{8cm}}
\hline
Ref. & Dataset & \# of Ents.\&Rel.& \makecell{Joint Technique \& \\ F1 Scores} & Limitations \\
\hline
Guo et al. 2021~\cite{CyberRel} & Custom & 8, 7 &\makecell{BERT-BiGRU-CRF\\ F1 = 81\% } & Tagging representation loses the semantic relevance of entity types; Model complexity and low performance scores   \\

\hline
Zuo et al. 2022\cite{ZuoYali} & Custom & 7, 6 & \makecell{BERT-BiLSTM-CRF\\ F1 = 70\%} & The use of a separate relation matching rule to enhance the relation accuracy, which contradicts the fundamental principle of JE; Low F1 scores;  \\
\hline
Ahmed et al. 2024~\cite{CyberEntRel} & Custom & 12, 11& \makecell{RoBERTa-BiGRU-CRF \\ F1 = 83\%}  & The use of a separate relation matching rule to enhance the relation accuracy, which contradicts the fundamental principle of JE. \\

\hline
You et al. 2024~\cite{TiGNet} & Custom & -, 7& \makecell{Chinese\_BERT-BiGRU-FC \\ F1 = 71\%} & Tailored for Chinese texts; Focuses on RE rather than comprehensively addressing both entities and relations; Low F1 scores \\

\hline
Liu et al. 2024~\cite{CTI-JE} & Custom & 9, 5& \makecell{SecBERT-BiLSTM-DCNN\\ F1 = 74\%} & Multiple layers architecture with low precision, recall, and F1 scores. \\

\hline
Wang et al. 2024~\cite{XiaodiWang} & Custom & 9, 6& \makecell{ BERT-DCNN-Att \\ F1 = 87\%} & Requires manually converting relation types to semantically unambiguous words; Architecture complexity and mitigated performance. \\
\hline
Lv et al. 2024~\cite{CTI-TFN} & Custom & 9, 5 &\makecell{BERT-BiLSTM + Fnets \\NER F1= 88\% \\ Av. RE F1 = 78\%} & Ignores complex relation types and overlapping relation challenges; low precision, recall, and F1 scores.\\
\hline
 Zhu et al. 2024~\cite{ITIRel} & Custom & 13, 6 &\makecell{ BERT + Token-pair Matrix \\NER F1= 92\% \\ Av. RE F1 = 88\%} & High computation complexity; Ignore complex relation types; Uses a static dictionary.\\
\hline
\textbf{TIJERE} & \textbf{Public} & \textbf{13, 15}& \makecell{\textbf{Data-driven Approach with}\\ \textbf{EDF + MSLR} \\ \textbf{NER: F1 = 93\%} \\ \textbf{Triple RE: F1 = 98}} & \textbf{Requires technical data modeling, transformation, and implementation} \\

\bottomrule
\end{tabular}
\label{tab:je_related_wrok}
\end{table*}
Miwa and Bansal~\cite{MakotoMiwa} implemented a multi-task sequential learning approach that stacks bidirectional tree-structured and sequential LSTM-RNNs to jointly extract entities and relations from news articles. Their model outperformed baseline models, including those that use convolutional neural networks (CNNs) on public datasets (ACE04, ACE05, and SemEval-2010). Bekoulis et al.~\cite{GiannisBekoulis} designed a joint approach that incorporates CRF for entity recognition and a multi-headed selection method for RE. Its effectiveness was demonstrated by applying the model to open-source news datasets, including ACE04, CoNLL04, and DREC. Wang et al.~\cite{BERT-CRF} proposed an approach using BERT-CEF to extract head entities and BiLSTM to simultaneously identify the tail entity relation, and address overlapping relation challenges in English and Chinese datasets. Similarly, Yuan et al.~\cite{YueYuan} introduced RSAN, a relation-based attention network that leverages the BiLSTM architecture. Specifically, the authors designed relation-specific sentence representations and employed sequence labeling to extract head and tail entities for each relation from the text.\\
Although these studies implement the JE approach, they predominantly utilize traditional machine learning (ML) methods and focus on extracting entities such as persons, organizations, and locations from news articles. These entities typically exist within well-defined contexts and have relatively clear semantic categories. Contrastingly, the cybersecurity forensic datasets used in this study involve entities related to threats, vulnerabilities, and attacks, presenting higher complexity and ambiguity. These datasets include non-standard terms (e.g., Mimikatz) that lack definitions outside of the cybersecurity domain. Consequently, they demand advanced NLP techniques, such as transformer models and specialized data representations for accurate feature extraction and interpretation.

The use of the JE approach is still in the early stages of adoption in cybersecurity. Guo et al.~\cite{CyberRel} proposed CyberRel based on the pretrained BERT model, to simultaneously extract entity-relation triples from unstructured security reports. More specifically, their approach was formulated as a multiple sequence labeling task using a tagging scheme where each relation has its sentence representation containing subject-object entities. However, a considerable problem with this approach is its oversight of the relevance of entity categories in the tagging scheme, which is crucial to enable the interaction between sub-tasks and improve relation classification accuracy. Consequently, despite stacking both an encoder and a decoder, the model demonstrates relatively low performance with an F1-score of 0.81 on a dataset of only 8 entities and 7 relation types.\\
Other related works include Zuo et al.~\cite{ZuoYali}, who proposed an end-to-end attention-based model utilizing a novel tagging scheme to jointly extract entities and their relationships from security texts. Similarly, Ahmed et al.~\cite{CyberEntRel} introduced RoBERTa-BiGRU-CRF, an attention-based model that demonstrates superior performance compared to other state-of-the-art methods, including the model developed by Zuo et al.~\cite{ZuoYali}. However, while these studies adopt tagging representations that effectively handle overlapping relations in complex sentences, they lack sufficient semantic contextual information to differentiate task-specific features and improve classification performance. Consequently, they incorporate separate relation-matching rules to mitigate feature confusion and enhance RE accuracy. This reliance on an additional module for relation classification makes their approach resemble a PE method, which contradicts the core principles of JE.
You et al.~\cite{TiGNet} used token-pair matrices to convert JE into multiple token-span recognition tasks. Their TiGNet framework integrates token position features to enhance classification results. However, TiGNet is tailored for Chinese texts and primarily focuses on RE rather than comprehensively addressing both entities and relations. 
 Liu et al.~\cite{CTI-JE} deployed CTI-JE, which models JE as a table-filling problem using task-specific representation. Despite leveraging task-specific representation and employing dilated convolution (DConv) for feature extraction, CTI-JE exhibited relatively low performance in terms of precision, recall, and F1 scores.
Lv et al.~\cite{CTI-TFN} developed CTI-TFN, a task-specific Fourier network-based model for joint entity and RE from CTI reports. Their model utilizes two independently weighted Fourier networks (Fnets) based on BiLSTM, with one dedicated to entity recognition and the other to RE, ensuring specialized feature learning for each task. However, the approach overlooks challenges such as overlapping relations and complex relation types, which limit the adaptability of the model and generalization. Zhu et al.~\cite{ITIRel} proposed ITIRel, a joint entity and RE framework for Internet-of-Things threat intelligence that leverages domain knowledge and token-pair representations to address overlapping challenges and enhance performance. However, its reliance on a static dictionary restricts adaptability to new or emerging patterns. Additionally, constructing an $n \times n$ token-pair matrix (where n indicates the number of tokens in a sentence) significantly increases computational complexity, resulting in training times exceeding half a day, a problem that worsens with larger, more complex datasets. Contrastingly, our proposed TIJERE model adopts a data-driven approach and avoids static dictionaries by introducing an innovative MSLR approach. MSLR encodes entity-centric features to enhance positional and contextual understanding, enabling accurate extraction using a simpler and more efficient architecture. ITIRel achieved comparable results to our proposed model for NER, largely owing to the nature of the dataset used by Zhu et al., which focuses on product and vulnerability descriptions. These texts are generally less complex and follow recognizable patterns such as operating systems, hardware, and CVE identifiers, which are easier for ML models to capture. By contrast, the APT reports used in this work present greater complexity and ambiguity, requiring models to leverage contextual information to distinguish between nuanced attack techniques and behaviors of APT groups. This distinction was also demonstrated in our previous studies~\cite{TiKG,CTiKG}.\\ 
 Wang et al.~\cite{XiaodiWang} aimed to enhance JE accuracy by introducing the enhanced representation and binary tagging framework, which concatenates relation embedding representation and word embedding at the embedding layer to enhance RE. Despite the approach being innovative, their model demonstrates only comparable or slightly improved performance relative to existing methods in terms of precision, recall, and F1 scores. Furthermore, they did not adopt any entity encoding schema in their dataset annotation, which resulted in the implementation of two binary taggers to identify subject and object entities, adding an extra layer of complexity to the model. Additionally, replacing relation types with unambiguous words to enrich feature representations is subjective and prone to errors, ultimately hindering the generalization ability of the model.\\
This paper introduces TIJERE, a data-driven framework to jointly extract entities and their interconnections from threat intelligence reports. TIJERE employs MSLR to reinforce entity recognition and effectively address the overlapping relations. It further integrates data-centric features such as entity mask and entity type, which inject semantic constraints and contextual insights, facilitating interaction between sub-tasks and enabling the model to make more informed decisions, thereby improving RE accuracy. Additionally, TIJERE leverages SecureBERT$^+$, an enhanced version of SecureBERT~\cite{SecureBERT}, to introduce domain-specific security context and reduce the inherent ambiguity in cybersecurity data. The framework demonstrates significant performance gains on a large dataset featuring diverse entities and relations, highlighting its effectiveness in real-world cybersecurity applications. \\
Table~\ref{tab:je_related_wrok} summarizes relevant JE studies in threat intelligence. It indicates that the proposed model significantly outperforms existing approaches, even when evaluated on a scalable dataset with a large number of entities and relations, highlighting its robustness and effectiveness in real-world scenarios.

\section{TIJERE Framework}
This section provides a detailed overview of the TIJERE framework. As a data-driven approach, the methodology begins with the dataset presentation and its structure, followed by expert feature engineering, MSLR, model architecture, and implementation details. This structured presentation ensures a smooth transition for readers, offering a clear understanding of the design and functionality of the framework.
\subsection{Dataset}
This study extends the open-source DNRTI\cite{DNRTI} dataset, as released by~\cite{TiKG}. The dataset comprises 175,220 tokens and includes 36,808 annotated entities, all labeled by threat intelligence experts using the BIO tagging format. In this format, \textbf{B} (Beginning) denotes the start of an entity, \textbf{I} (Inside) indicates that the token is part of an entity, and \textbf{O} (Outside) signifies that the token does not belong to any entity. The dataset encompasses 12 entity types, including HackOrg, OffAct, SamFile, SecTeam, Tool, Time, Purp, Area, Org, Way, Exp, and Features. Table \ref{tab:datasetInfo} summarizes the distribution of entity categories in the dataset, as described by~\cite{TiKG}.
\begin{table}[h]
\centering
\caption{Distribution of entity types in the dataset.}
\begin{tabular}{l|p{2cm}|p{2.3cm}|p{.7cm}}
\hline
\textbf{Entity types} & \textbf{Meaning} &\textbf{Examples} &\textbf{Count} \\ \hline
Area     & Location & China & 3435 \\\hline
Exp      & Exploit & EternalBlue & 1995 \\\hline
Features & Characteristics & Collect information & 2442 \\\hline
HackOrg  & Hacker group&MuddyWater & 4490 \\\hline
OffAct   & Attack pattern & Brute force & 2682 \\\hline
Org      & Industry & Military industry & 4696 \\\hline
Purp     & motivation & DoS & 2458 \\\hline
SamFile  & File& Worksheets & 2341 \\\hline
SecTeam  & Security team & FireEye & 1920 \\\hline
Time     & Time & April 20, 2016 & 2657 \\\hline
Tool     & Tool & Clayslide, LaZagne & 4917 \\\hline
Way      & Technique & Spear-phishing & 2037 \\ \hline
\end{tabular}
\label{tab:datasetInfo}
\end{table}
The dataset was chosen for its public availability and diverse entity categories, enabling the definition of multiple relation types for model evaluation. It describes APT attacks and includes complex entities such as "Features", "Purp", and "Way", which require contextual understanding for accurate classification. This makes it more representative of real-world challenges compared to datasets like STUCCO and CyNER, which primarily contain vulnerability descriptions with more straightforward pattern recognition for ML models~\cite{TiKG}.  
\begin{figure}[htbp]
\centerline{\includegraphics[width=.5\textwidth, height=3in]{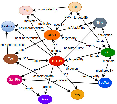}}
\caption{A sample schema showing entity relationships within the dataset}
\label{fig:ontology}
\end{figure}
The DNRTI dataset was originally released for the NER task. To extend it for RE, Mouiche and Saad~\cite{TiKG} predefined 15 relation types and annotated the dataset to evaluate the RE module of their pipeline framework. They developed a domain-specific ontology schema, Figure~\ref{fig:ontology}, to characterize and reveal entity interactions within the dataset. UCO~\cite{UCO} provides a comprehensive definition of each relation type, specifying its Domain and Range:
\begin{itemize}
    \item analyses: Domain:secTeam\\
    Range:SamFile
    \item associatedWith: Domain:HackOrg.\\
    Range: HackOrg
    \item discovers: Domain:SecTeam\\
    Range: HackOrg
    \item discoveredBy: Domain:HackOrg\\
    Range:SecTeam
    \item hasAttackTime: Domain:HackOrg or OffAct or Way\\
    Range:Time
    \item hasCharacteristics: Domain:HackOrg or OffAct or Exp or Way or Tool or SamFile.\\
    Range:Features
    \item locatedAt: Domain:Org
    Range:Area
    \item monitors: Domain:SecTeam\\
    Range:Org or Area or Tool or Exp
    \item monitoredBy: Domain:Org or Area or Tool or Exp\\
    Range:SecTeam
    \item motivates: Domain:Purp\\
    Range:HackOrg or OffAct or Exp or Way 
    \item motivatedBy: Domain:HackOrg or OffAct or Exp or Way.\\
    Range:Purp
    \item uses: Domain:HackOrg or OffAct or Exp or Way or Tool or SamFile.\\
    Range:Tool or OffAct or Exp or SamFile or Way
    \item usedBy: Domain:Feaures or OffAct or Exp or Way or Tool or SamFile.\\
    Range:HackOrg or OffAct or Exp or Way or Tool or SamFile
    \item targets: Domain:HackOrg or OffAct or Exp or Way or Tool or SamFile.\\
    Range:Area or Org or SecTeam
    \item targetedBy: Domain:Area or Org or SecTeam\\
    Range:HackOrg or OffAct or Exp or Way or Tool or SamFile
\end{itemize}

\begin{figure}[htbp]
\centerline{\includegraphics[width=.5\textwidth, height=1.8in]{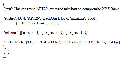}}
\caption{Sample data format in the DNRTI-JE dataset for joint extraction.}
\label{fig:data_format}
\end{figure}

We adopted the same relation definitions and methodology and introducd the "noRelation" type to consider entity pairs that do not have a defined relationship in the schema (e.g., between "Time" and "Org" entities). Unlike~\cite{TiKG}, which maintains separate annotated datasets for each subtask, we restructured the dataset into a unified JSON format suitable for JE. The resulting dataset, DNRTI-JE, was carefully cross-validated by the authors to ensure annotation accuracy, consistency across entity and relation labels, and alignment with the defined ontology. It will be made publicly available on GitHub to support reproducibility and facilitate further research. A sample data record from DNRTI-JE is presented in Figure~\ref{fig:data_format}. Each data sample contains four primary variables or components:
\begin{itemize} 
\item \textbf{text}: A sentence extracted from CTI reports describing an attack scenario. 
\item \textbf{entities}: A list of annotated entities within the "text," including their respective start and end indices within the sentence. 
\item \textbf{relations}: A list of identified relations for each valid entity-pair in the "entities" list. The indices represent the position of the head and tail entities in the "entities" list.
\item \textbf{entity\_labels}: A list of labels assigned to each token in the sentence. \end{itemize}
\begin{table}[h]
\centering
\caption{Relation distribution in the DNRTI-JE dataset.}
\begin{tabular}{l|c}
\hline
\textbf{Relation Type} & \textbf{Count}\\ \hline
associatedWith   & 2057\\ \hline
analyses  & 1723\\ \hline
discovers  & 2114\\ \hline
discoveredBy & 1061\\ \hline
hasAttackTime & 1810\\ \hline
hasCharacteristics & 2087 \\ \hline
locatedAt & 2258\\ \hline
monitors & 2130\\ \hline
monitoredBy & 1250\\ \hline
motivates & 1620\\ \hline
motivatedBy & 2043\\ \hline
noRelation & 8062\\ \hline
targets & 4993\\ \hline
targetedBy & 2018\\ \hline
uses& 4114\\ \hline
usedBy & 3001\\ \hline
\end{tabular}
\label{tab:re_dist}
\end{table}
Table~\ref{tab:re_dist} presents the distribution of relation types in the DNRTI-JE dataset. As with many real-world datasets, it is imbalanced, which makes it suitable for evaluating the scalability and adaptability of our approach in complex scenarios. 

\subsection{Expert Feature Engineering}
As discussed in the Introduction, prior work has consistently shown that RE remains the most challenging task within the JE paradigm~\cite{XiaodiWang,CyberEntRel,ZuoYali}. This subsection presents the engineering of key expert-driven features designed specifically to support relation classification in joint settings. When integrated into the model, these features encode contextual, semantic, and positional information, thereby enhancing the ability of the model to distinguish between features and make accurate and informed predictions.
\begin{figure}[htbp]
\centerline{\includegraphics[width=.45\textwidth, height=.8in]{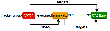}}
\caption{Graphical example of an annotated sentence containing overlapping relations.}
\label{fig:sample_ann_sent}
\end{figure}
An annotated example sentence is used, as illustrated in Figure~\ref{fig:sample_ann_sent}, to ensure a smooth transition in explaining how these features are extracted. This sentence contains three relation triples: (APT29, uses, Mimikatz), (APT29, targets, XYZ Bank), and (Mimikatz, targets, XYZ Bank). Considering the first relation triple (APT29, uses, Mimikatz), the challenge is determining what key information leads a security analyst to classify the relation as "uses" rather than one of the remaining 15 relation types. The decisive factor is the contextual and semantic understanding of APT29 and Mimikatz within the cybersecurity domain. More specifically, a security analyst understands that APT29 refers to a hacker group, while Mimikatz represents software or a tool used against the victim. This domain-specific knowledge naturally suggests that APT29 (HackOrg) is using Mimikatz (Tool), resulting in the correct "uses" classification. It appears that the entity types HackOrg (i.e., hacker organization) for APT29 and Tool (i.e., malware or software) for Mimikatz carry and can provide this crucial expert information to the model during training for accurate classification at inference. Without this contextual information, a model may misinterpret APT29 and Mimikatz, potentially associating APT29 with an apartment number and Mimikatz as an arbitrary or meaningless term. Since Mimikatz lacks a definition in general vocabulary outside of cybersecurity, this ambiguity is another source of feature confusion, making it challenging for ML models to accurately extract cybersecurity-related entities and relationships. This lack of domain awareness significantly impacts the ability of the model to differentiate between security-specific terms and unrelated text, resulting in inaccurate extraction and classification. The same reasoning applies to the other relation triples, demonstrating how entity type embeds domain-specific context. By incorporating entity type, the model leverages cybersecurity knowledge, ensuring relation classification is guided by expert-driven semantics rather than ambiguous textual patterns.

This study also introduces entity mask as another crucial feature for tracking the positions of entity pairs within a sentence. Entity mask is a vector-based representation, where tokens that belong to the entities involved in a given relation are marked with "1", while all other tokens are assigned "0". This mechanism provides precise positional information, enabling the model to focus on relevant segments of the sequence during training for the relation classification.
As depicted in Figure~\ref{fig:mslr}, entity mask highlights the tokens that correspond to the entity pair under consideration, ensuring that the model pays attention to the correct entities within complex sentence structures containing many relations. Entity mask reinforces the localization of entities and improves their contextual processing to enhance RE accuracy and strengthens the ability of the model to distinguish between overlapping and distant relations. Both entity type and entity mask as known as expert domain features (EDFs).
\subsection{Multisequence Labeling Representation}
\begin{figure*}[htbp]
\centerline{\includegraphics[width=.9\textwidth, height=2.8in]{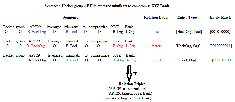}}
\caption{Multisequence labeling representation.}
\label{fig:mslr}
\end{figure*}
Upon identifying the determining factors, referred to as EDFs, that guide security analysts' annotation decisions, the following critical challenge remains: effectively conveying this domain-specific information to the ML model. To address this, we reformulate the JE task as a multisequence labeling problem and generate a separate sequence for each entity pair or relation within a sentence.
Figure~\ref{fig:mslr} illustrates the MSLR approach for an example sentence containing three entities and three relations. MSLR is composed of four key components: sequence, relation label, entity type, and entity mask, each designed to address major challenges associated with JE.
\newline\textbf{Sequence and Relation Label:}
A sequence is a full copy of the sentence, labeled specifically for a single relation. As the model needs to predict one relation per valid entity pair, multiple sequences are generated from a single sentence, one per relation. In the example shown in Figure~\ref{fig:sample_ann_sent}, this results in three distinct sequences for three identified relations. This structured formulation effectively resolves the challenge of overlapping relations, even in complex sentence structures.
\newline \textbf{Entity Type and Entity Mask:}
By observing only the sequence and relation label columns in Figure~\ref{fig:mslr}, it is evident that the model cannot rely solely on the generated sequences and relation types to extract meaningful classification features. The embedded input sequences are semantically identical but correspond to different relation categories, which results in feature confusion. The challenge is further augmented when the sentence lacks explicit relational cues or involves subtle relation types (e.g., hasAttackLocation), making it difficult for the model to distinguish between sequence features, resulting in suboptimal performance.
To overcome this, MSLR incorporates the identified expert-driven features: entity type and entity mask. Entity type encodes essential semantic and contextual information related to the entities involved in a relation. This helps refine the representation of each sequence during training and improves the ability of the model to interpret entity roles in context. Entity mask explicitly indicates the positions of the subject and object entities in the sequence. This guides the model to focus on the relevant token-level interactions when processing the sequence.
Collectively, these features empower the model to move beyond surface-level similarity and capture deeper relational semantics. For instance, during training, the model can learn that an interaction between a hacker organization (HackOrg) and a tool (Tool) is likely to indicate a "uses" relation. This enriched and structured representation enables the model to distinguish between otherwise ambiguous sequences, significantly enhancing its ability to classify subtle and complex relations with higher accuracy.
\newline \textbf{Language Ambiguity:} TIJERE addresses this challenge using SecureBERT$^+$, a domain-specific language model pretrained on cybersecurity texts. Compared to generic pretrained models, SecureBERT$^+$ can capture nuances and understand the jargon used in security reports, reducing entity misclassification and enhancing overall model performance. Additionally, MSLR inherently oversamples the dataset by duplicating input sentences for each entity-pair, which benefits the NER task by reinforcing entity recognition in diverse relational contexts.

References \cite{CyberRel,ZuoYali,CyberEntRel} employed relation-oriented representations to tackle overlapping relation challenges. However, these studies assign the label 'O' to all tokens that are not directly involved in a specific relation, potentially degrading the NER task. Contrastingly, our proposed approach preserves the original token labels while selectively highlighting entities relevant to each relation type. This strategy effectively handles overlapping relations and enhances NER by oversampling input sentences without losing token-level information. Furthermore, MSLR integrates expert-driven features, including entity mask to track entity pair positions and entity type to provide crucial contextual and semantic information, which results in improved relation classification accuracy.

\subsection{TIJERE Architecture}
\begin{figure*}[htbp]
\centerline{\includegraphics[width=.7\textwidth, height=4in]{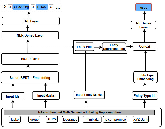}}
\caption{TIJERE Model's Architecture.}
\label{fig:tijere}
\end{figure*}
Figure~\ref{fig:tijere} illustrates the architecture of the proposed TIJERE system. TIJERE is composed of eight layers:
\subsubsection{Tokenization and Multisequence Labeling Representation}
Given an input data sample in JSON format, as depicted in Figure~\ref{fig:data_format}, this layer dynamically and automatically generates the MSLR. It processes each input sentence to produce the following structured inputs for every generated sequence:

\begin{itemize}
    \item $\mathbf{X} = { x_1, x_2, \dots, x_T }$ represents a set of tokens in a sequence of length $T$. This sequence undergoes tokenization to generate $\mathbf{I}_{\text{input\_ids}}$, which corresponds to the tokenized representation of the input sequence, ready for embedding and processing by the model.
    \item $\mathbf{I}_{\text{mask}} \in \{0, 1\}^T$ represents the input mask where each position $i$ is one if $x_i$ is a real token, and zero if it is padding.
    \item $\mathbf{I}_{\text{entity\_mask}} \in \{0, 1\}^T$ highlights the relevant parts of a sequence by focusing specifically on tokens involved in a given relation. Tokens that correspond to the entities participating in the relation are assigned a value of "1", while all other tokens are marked as "0", ensuring precise entity localization and enhancing relation classification accuracy (Figure~\ref{fig:mslr}). 
    \item $\mathbf{I}_{\text{entity\_type\_ids}} \in \mathbb{Z}^n$ represents the label IDs assigned to different entity types, where $n=11$ corresponds to the total number of entity categories (e.g., "HackOrg", "OffAct", etc.). This feature provides essential semantic context, enabling the model to understand entity roles and interactions within a given relation. The model incorporats entity type information to better distinguish relationships during training and inference. For example, recognizing one entity as a "HackOrg" and another as a "Tool" strongly indicates a "uses" relation, mimicking the analytical reasoning used by security experts.
\end{itemize}
\subsubsection{SecureBERT$^+$ Embedding}
The SecureBERT$^+$ embedding layer transforms each token $x_i$ into a cybersecurity contextualized embedding $\mathbf{h}_i^{\text{SecureBERT$^+$}} \in \mathbb{R}^{d_{\text{SecureBERT$^+$}}}$:
\begin{equation}
\begin{split}
    \mathbf{H}^{\text{SecureBERT$^+$}}& = \text{SecureBERT$^+$}(\mathbf{X}, \mathbf{I}_{\text{mask}}) \\
    &= \{ \mathbf{h}_1^{\text{SecureBERT$^+$}}, \mathbf{h}_2^{\text{SecureBERT$^+$}}, \dots, \mathbf{h}_T^{\text{SecureBERT$^+$}} \}
\end{split}
\end{equation}

where $\mathbf{H}^{\text{SecureBERT$^+$}} \in \mathbb{R}^{T \times d_{\text{SecureBERT$^+$}}}$.
This layer generates embeddings that capture both the semantic meaning of the token and the context in which it appears. These are known as dynamic embeddings, as they adapt based on the context of the token within the sentence.
\subsubsection{BiGRU Layer}

The BiGRU layer processes $\mathbf{H}^{\text{SecureBERT$^+$}}$ to capture sequential dependencies:
\begin{equation}
    \mathbf{H}^{\text{BiGRU}} = \text{BiGRU}(\mathbf{H}^{\text{RoBERTa}})
\end{equation}

where $\mathbf{H}^{\text{BiGRU}} = \{ \mathbf{h}_1^{\text{BiGRU}}, \mathbf{h}_2^{\text{BiGRU}}, \dots, \mathbf{h}_T^{\text{BiGRU}} \} \in \mathbb{R}^{T \times 2 \cdot d_{\text{hidden}}}$ and $d_{\text{hidden}}$ represents the hidden dimension for the GRU in each direction.

\subsubsection{Entity Type Embedding Layer}

Each token’s entity type ID $\mathbf{I}_{\text{entity\_type}}$ is embedded as:
\begin{equation}
    \mathbf{E}_{\text{entity}} = \text{Embedding}(\mathbf{I}_{\text{entity\_type}})
\end{equation}

where $\mathbf{E}_{\text{entity}} \in \mathbb{R}^{T \times d_{\text{entity}}}$, with $d_{\text{entity}}$ matching $2 \cdot d_{\text{hidden}}$ (BiGRU output dimension). Unlike the SecureBERT$^+$ layer, entity\_type embedding is a static representation of the entity type, independent of the context. If the entity pair is ("APT29", "Mimikatz"), the entity type embedding will encode "HackOrg" and "Tool", respectively, providing explicit entity type information to the model during relation classification.

\subsubsection{NER Classification Layer (NER Dense and CRF Layers)}

The NER classifier predicts entity tags for each token based on $\mathbf{H}^{\text{BiGRU}}$:
\begin{equation}
    \mathbf{z}_{\text{NER}} = \text{Dense}(\mathbf{H}^{\text{BiGRU}})
\end{equation}
where $\mathbf{z}_{\text{NER}} \in \mathbb{R}^{T \times L_{\text{NER}}}$ and $L_{\text{NER}}$ indicates the number of NER label classes.\\

The CRF layer then takes $\mathbf{z}_{\text{NER}}$ to output the predicted sequence of entity labels:
\begin{equation}
    \hat{\mathbf{y}}_{\text{NER}} = \text{CRF}(\mathbf{z}_{\text{NER}})
\end{equation}

\subsubsection{Entity Pooling and Representation (for RE)}

The entity pooling operation aggregates token embeddings using $\mathbf{I}_{\text{entity\_mask}}$, producing a combined representation of the subject and object entities:
\begin{equation}
    \mathbf{e}_{\text{pool}} = \sum_{i=1}^{T} \mathbf{h}_i^{\text{BiGRU}} \cdot \mathbf{I}_{\text{entity\_mask}, i} \label{eq:pooling}
\end{equation}

This yields the entity representation, which is concatenated with the entity type embedding:
\begin{equation}
 \mathbf{e}_{\text{concat}} = \text{Concat}(\mathbf{e}_{\text{pool}}, \mathbf{E}_{\text{entity}})
\end{equation}
Entity type embedding encodes the types of the two entities.
\subsubsection{Relation Classification (RE Dense Layer)}

The concatenated entity representation $\mathbf{e}_{\text{concat}}$ is then passed through a dense layer for relation classification:
\begin{equation}
  \mathbf{z}_{\text{RE}} = \text{Dense}(\mathbf{e}_{\text{concat}})
\end{equation}
The final relation prediction is given by applying Softmax over $\mathbf{z}_{\text{RE}}$:
\begin{equation}
    \hat{\mathbf{y}}_{\text{RE}} = \text{softmax}(\mathbf{z}_{\text{RE}})
\end{equation}
By combining the entity types with the pooled hidden states from the sentence, the RE classifier utilizes these expert-crafted features to predict the relationship between the two entities with greater precision.

\subsubsection*{Total Loss Calculation}
The total loss for the joint model during training combines the NER loss (CRF loss) and RE loss (cross-entropy loss) with weights $\alpha$ and $\beta$:
\begin{equation}
   \mathcal{L}_{\text{Total}} = \alpha \cdot \mathcal{L}_{\text{NER}} + \beta \cdot \mathcal{L}_{\text{RE}}
   \label{eq:loss}
\end{equation}
where:
\begin{align*}
    \mathcal{L}_{\text{NER}} &= -\log P(\hat{\mathbf{y}}_{\text{NER}} | \mathbf{z}_{\text{NER}}) \\
    \mathcal{L}_{\text{RE}} &= -\sum_{c} y_{\text{RE}}^{(c)} \log \hat{\mathbf{y}}_{\text{RE}}^{(c)}
\end{align*}

The total loss in Eq.~\ref{eq:loss} captures the joint optimization of the NER and RE sub-tasks. During backpropagation, this combined loss is reinjected into the shared encoder layers, enabling mutual learning, reducing error propagation, and improving the overall performance of the model. In our implementation, the weighting coefficients are set to $\alpha = \beta = 1$ to balance the contribution of both objectives.

\subsection{Experiment and Evaluation}
In this section, the implementation of TIJERE is presented and its performance is compared with state-of-the-art JE models in cybersecurity using the DNRTI-JE dataset.

\subsubsection{Training and Evaluation}
To train and evaluate the TIJERE model, the DNRTI-JE dataset was divided into training, validation, and testing sets with a split ratio of 70\%, 15\%, and 15\%, respectively. Table  \ref{tab:modelparams} lists the base model parameter settings and GPU information for these experiments.
\begin{table}[h]
\centering
\caption{Model parameter settings.}
\begin{tabular}{l|c}
\hline
\textbf{Parameters} & \textbf{Values}\\ \hline
batch size    & 16 \\ \hline
dropout  & 0.3 \\ \hline
epsilon & 1e-8 \\ \hline
initial learning rate & 1e-5\\ \hline
hidden layer size & 256 $\times$ 2 \\ \hline
embedding size & 768 \\ \hline
number of epochs & 3 \\ \hline
optimizer & AdamW \\ \hline
kernel\_size & 3 \\ \hline
weight\_decay & 0.01 \\ \hline
conv\_layer & 4 \\ \hline 
\multicolumn{2}{|c|}{\textbf{GPU Information}} \\ \hline
 GPU type &  NVIDIA Tesla T4\\ \hline
 CUDA Version & 12.1 \\ \hline
 TensorFlow version & 2.17.0 \\ \hline
 Torch version & 2.5.0+cu121 \\ \hline
 Mem allocated & 3.8GB\\ \hline
 Mem cached & 6GB \\ \hline
\end{tabular}
\label{tab:modelparams}
\end{table}

\begin{table*}[h]
\centering
\caption{TIJERE evaluation results and comparisons. EDF: Expert domain features; DCNN: Dilated convolutional neural networks.}
\begin{tabular}{p{3.2cm}|p{.37cm}|p{.37cm}|p{.37cm}|p{.55cm}|p{.37cm}|p{.37cm}|p{.37cm}|p{.6cm}||p{.4cm}|p{.4cm}|p{.4cm}|p{.55cm}|p{.4cm}|p{.4cm}|p{.4cm}|p{.55cm}}
\hline
 & \multicolumn{8}{c||}{\textbf{EDF Disabled}} & \multicolumn{8}{|c}{\textbf{EDF Enabled}} \\
 
\cline{2-9} \cline{10-17} 
 & \multicolumn{4}{c}{\textbf{NER}} & \multicolumn{4}{|c||}{\textbf{RE}} & \multicolumn{4}{|c|}{\textbf{NER}} & \multicolumn{4}{|c}{\textbf{RE}} \\
    \cline{2-5}  \cline{6-9} \cline{10-13} \cline{14-17}

\textbf{Models} & \textbf{P} &\textbf{R} &\textbf{F1} & \textbf{Acc} &\textbf{P} &\textbf{R}& \textbf{F1} &\textbf{Acc} &\textbf{P} & \textbf{R} & \textbf{F1}& \textbf{Acc} & \textbf{P}& \textbf{R}& \textbf{F1}& \textbf{Acc}\\ \hline
BERT-CRF~\cite{BERT-CRF} & 0.83 & 0.84 & 0.83 & 0.995 & \textbf{0.51} & \textbf{0.42} & \textbf{0.44} & \textbf{0.550} & 0.82 &0.86 & 0.84 & 0.995 & \textbf{0.98} & \textbf{0.97} & \textbf{0.97}& \textbf{0.984}\\ \hline

BERT-BiGRU-CRF~\cite{CyberRel} & 0.85 & 0.86 & 0.86 & 0.991  & \textbf{0.48}& \textbf{0.41} & \textbf{0.42} &\textbf{0.511} & 0.86 & 0.86& 0.86 & 0.991& \textbf{0.96} &\textbf{0.97}  &\textbf{0.97} & \textbf{0.971}\\ \hline
BERT-BiLSTM-CRF \cite{Vulcan,ZuoYali} & 0.85 & 0.86 & 0.85 & 0.989  & \textbf{0.50}& \textbf{0.45} & \textbf{0.48} &\textbf{0.526} & 0.84 & 0.85& 0.85 & 0.991& \textbf{0.97} &\textbf{0.97}  &\textbf{0.97} & \textbf{0.972}\\ \hline

BERT-DCNN-CRF~\cite{XiaodiWang} & 0.84 & 0.86 & 0.85 & 0.991 & \textbf{0.44}& \textbf{0.47} & \textbf{0.46} & \textbf{0.486}& 0.84 &0.86 & 0.85 &0.993 & \textbf{0.95} & \textbf{0.97} & \textbf{0.96} & \textbf{0.974}\\ \hline

RoBERTa-BiGRU-CRF~\cite{CyberEntRel,OpenCyKG} & 0.89 & 0.92  & 0.90 & 0.991 & \textbf{0.50} & \textbf{0.40} & \textbf{0.45} & \textbf{0.508}& 0.90 & 0.91 & 0.90 & 0.992 & \textbf{0.97}  & \textbf{0.98} & \textbf{0.97} & \textbf{0.971}\\ \hline
SecureBERT-DCNN-CRF & 0.90 & 0.91 & 0.90 & 0.983 &\textbf{0.43} &\textbf{0.43}  &\textbf{0.43} & \textbf{0.475}& 0.90 &0.90 & 0.90 & 0.986  & \textbf{0.95} & \textbf{0.98} &\textbf{0.95} & \textbf{0.972}\\ \hline
SecureBERT-BiGRU-CRF~\cite{CTiKG} & 0.92 & 0.92 &0.91  & 0.993 & \textbf{0.47} & \textbf{0.46} &\textbf{0.45} & \textbf{0.495} & 0.91 &0.93 & 0.92 &0.995 & \textbf{0.98} & \textbf{0.98} & \textbf{0.98}& \textbf{0.971}\\ \hline
SecureBERT$^+$-BiGRU-CRF & 0.93 & 0.93 &0.93  & 0.994 & \textbf{0.45} & \textbf{0.46} &\textbf{0.44} & \textbf{0.495} & 0.93 &0.94 & 0.93 &0.995 & \textbf{0.98} & \textbf{0.99} & \textbf{0.98}& \textbf{0.976}\\ \hline

\hline
\end{tabular}
\label{tab:je_eval}
\end{table*}
To assess the effectiveness of the TIJERE model in addressing challenges in JE settings, we conducted an ablation study in two phases: first, without incorporating EDF (EDF disabled), and second, with EDF enabled. As the baseline for comparison, we implemented state-of-the-art approaches reported in Table~\ref{tab:je_related_wrok} from scratch, which have demonstrated significant performance in recent studies. The models based on SecureBERT and SecureBERT$^+$ are integral components of the proposed TIJERE framework. For NER, an entity is considered correctly extracted only if both its span and type are accurately identified. Similarly, for the RE task, a relationship is correctly extracted when both the entity pair and the corresponding relation type are correctly identified. The evaluation was conducted using Precision (P), Recall (R), and F1 score (F1) as key performance metrics.

The comparative results reported in Table \ref{tab:je_eval} reveal the followings:

\textbf{Effect of Expert Domain Features (EDF) on RE}: Incorporating EDF significantly improves RE performance across all models. For instance, BERT-CRF's RE F1 score increases from 0.44 to 0.97 when EDF is applied, highlighting the importance of expert-driven entity context. The impact of EDF is particularly evident in RE tasks, where it results in substantial accuracy and F1 score improvements even with a simplified architecture such as BERT-CRF, whereas NER performance remains relatively stable regardless of the use of EDF. The "EDF Disabled" corresponds to excluding the entity type and entity mask columns from Figure~\ref{fig:mslr}, reverting to a model-centric approach used in prior studies. This approach relies solely on the model to learn semantic representations and extract sentence features for both NER and RE. However, this strategy proves less effective, resulting in feature confusion, particularly for RE since identical sentences map to different relation types without explicit contextual and semantic differentiation. Additionally, some relation types (e.g., "hasAttackTime") provide small semantic signals, making it even harder for the model to distinguish between relevant patterns without explicit entity cues. This ultimately leads to lower classification accuracy, as demonstrated in the evaluation results in Table \ref{tab:je_eval}. 

  \textbf{Model Performance on NER Task}: The best performing models for NER, both with and without EDF enabled, are SecureBERT variants, which achieve F1 scores between 0.91 and 0.93, with accuracy reaching up to 0.993. Models that utilizes SecureBERT variants and RoBERTa consistently outperform standard BERT-based models, demonstrating the effectiveness of SecureBERT’s pretraining in capturing security-specific entity representations. Notably, while NER performance remains stable across different configurations, MSLR reinforces entity recognition by duplicating sentences while preserving their original labels. This oversampling effect enhances model training by improving the robustness and generalization of entity recognition while maintaining classification accuracy. However, it is at the expense of increased execution time.
   Moreover, our findings indicate that the 9\% improvement of SecureBERT$^+$ over the original SecureBERT in masked language modeling~\cite{SecureBERTPlus} tasks does not translate into entity recognition and RE tasks, where both models achieve similar performance.
   
    \textbf{Comparative Performance of Model Architectures}: Models that incorporate BiGRU layers, such as BERT-BiGRU-CRF and SecureBERT-BiGRU-CRF, generally outperform DCNN-based models. Although DCNN provides advantages in terms of speed and efficiency~\cite{XiaodiWang}, the NER task requires effective capturing of sequential relationships and contextual information. BiGRU excels in this regard due to its ability to process entire sequences in both directions, thereby enhancing contextual understanding and enhancing the performance of the model.
    
    \textbf{Execution Time}: The models have execution times ranging from approximately 72 to 84 min. This indicates moderate variability in processing times across different architectures. The BERT-CRF model has the shortest execution time at approximately 72 min owing to its simpler architecture. While adding layers such as BiGRU, BiLSTM, and DCNN or using enhanced Transformers such as SecureBERT improves model performance, it is achieved at the expense of increased computational costs.
Overall, the data-centric approach proposed in this paper effectively addresses the prevailing challenges in RE and demonstrates significant performance improvements over existing JE techniques in the cybersecurity domain. The use of SecureBERT embeddings markedly enhances the NER task, while RE benefits from the integration of expert-driven domain features. In essence, SecureBERT-based models consistently achieve the highest F1 scores for joint NER and RE tasks with EDF enabled, ensuring accurate extraction of entity-relation triples and outperforming baseline models in terms of both precision and recall.

\begin{figure*}[htbp]
\centerline{\includegraphics[width=.85\textwidth, height=3.in]{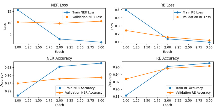}}
\caption{Training and validation losses and accuracies of SecureBERT$^+$-BiGRU-CRF, demonstrating its convergence.}
\label{fig:je_eval_fig}
\end{figure*}
Figure \ref{fig:je_eval_fig} further illustrates the performance of SecureBERT$^+$-BiGRU-CRF, showing training and validation losses along with accuracies over 4 epochs for both NER and RE tasks.
\begin{itemize}
    \item \textbf{NER Loss (top-left)}: The training NER loss exhibits a marked decrease over the epochs, starting above 15 and dropping below 2 by the third epoch, indicating effective learning on the training data. Contrastingly, the validation NER loss remains relatively stable, showing only a slight decline. This stability suggests that the model effectively generalizes without major overfitting. The small generalization gap is primarily owing to the difficulty in distinguishing between certain entities such as Features, OffAct, Purp, and Way, that depend heavily on contextual information for accurate classification.
    \item \textbf{RE Loss (top-right)}: Both the training and validation RE losses exhibit a steady decline over the epochs. Initially, the training loss drops sharply before stabilizing at a lower level, while the validation loss also decreases consistently. This trend indicates that the model can effectively learn to improve RE performance. The parallel reduction in both training and validation losses suggests good generalization, with minimal risk of overfitting for the RE task.
    \item \textbf{NER Accuracy (bottom-left)}: The accuracy of training for the NER task rises significantly throughout the epochs, nearing 0.99 by the end of the third epoch. Meanwhile, the validation NER accuracy also improves, albeit at a more gradual pace, leveling off at approximately 0.95. This slower rate of increase in validation accuracy relative to the training accuracy suggests that the model could be further improved in terms of generalization as mentioned above.
    \item \textbf{RE Accuracy (bottom-right)}: The RE accuracy on both training and validation data improves consistently over the epochs. By the third epoch, the validation accuracy stabilizes slightly below the training accuracy, with both metrics above 0.94. The increase in RE accuracy across both training and validation sets demonstrates the ability of the model to learn and generalize RE effectively. The close alignment between training and validation accuracies suggests low overfitting. 
\end{itemize}
\begin{figure}[htbp]
\centerline{\includegraphics[width=.45\textwidth, height=2.6in]{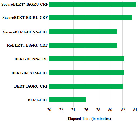}}
\caption{Model execution time.}
\label{fig:elapsed_time}
\end{figure}

These results confirm that the SecureBERT$^+$-BiGRU-CRF model is highly effective for both NER and RE tasks, demonstrating strong learning capabilities, convergence, and generalization. Figure \ref{fig:je_eval_fig} further validates the results presented in Table~\ref{tab:je_eval}, providing visual evidence of the performance and stability of the model.

\subsubsection{Ablation Study on Expert Domain Features for Relation Extraction}
\begin{table*}[h]
\centering
\caption{Ablation study results on expert domain features (EDF) for RE in joint settings. DCNN refers to Dilated CNNs.}
\begin{tabular}{p{3.6cm}|p{.4cm}|p{.4cm}|p{.4cm}|p{.4cm}|p{.4cm}|p{.45cm}|p{.4cm}|p{.4cm}|p{.4cm}|p{.4cm}|p{.4cm}|p{.4cm}|p{.4cm}|p{.4cm}|p{.45cm}}
\hline
    & \multicolumn{3}{c}{\textbf{NER}} & \multicolumn{12}{|c}{\textbf{RE}} \\
 
   \cline{5-16} 
 & \multicolumn{3}{c|}{\textbf{}} & \multicolumn{3}{|c|}{\textbf{EDF Disabled}} & \multicolumn{3}{|c|}{\textbf{Entity Mask Only}} & \multicolumn{3}{|c|}{\textbf{Entity Type Only}} &  \multicolumn{3}{|c}{\textbf{EDF Enabled}} \\
    \cline{2-4}  \cline{5-8} \cline{9-12} \cline{13-16}

\textbf{Models} & \textbf{P} &\textbf{R} &\textbf{F1} & \textbf{P} &\textbf{R} &\textbf{F1}& \textbf{P} &\textbf{R} &\textbf{F1} & \textbf{P} & \textbf{R}& \textbf{F1} & \textbf{P}& \textbf{R}& \textbf{F1} \\ \hline
BERT-CRF~\cite{BERT-CRF} & 0.82 & 0.83 & 0.82 & \textbf{0.48} & \textbf{0.51} & \textbf{0.46} & 0.82 & 0.82 & 0.82 &0.98 & 0.99 & 0.98 & \textbf{0.98} & \textbf{0.97} & \textbf{0.98}\\ \hline

BERT-BiGRU-CRF~\cite{CyberRel} & 0.86 & 0.86 & 0.86 & \textbf{0.51} & \textbf{0.48}& \textbf{0.41} & 0.78 & 0.80 & 0.79 & 0.97& 0.97 & 0.97& \textbf{0.97} &\textbf{0.97}  &\textbf{0.98} \\ \hline
BERT-BiLSTM-CRF \cite{Vulcan,ZuoYali} & 0.86 & 0.86 & 0.85 & \textbf{0.49}  & \textbf{0.45}& \textbf{0.45} & 0.78 & 0.79 & 0.79 & 0.97& 0.97 & 0.97& \textbf{0.97} &\textbf{0.98}  &\textbf{0.97} \\ \hline

BERT-DCNN-CRF~\cite{XiaodiWang} & 0.85 & 0.86 & 0.85 & \textbf{0.43} & \textbf{0.44}& \textbf{0.40} & 0.79 & 0.78& 0.78 &0.96 & 0.97 &0.97 & \textbf{0.96} & \textbf{0.97} & \textbf{0.96} \\ \hline

RoBERTa-BiGRU-CRF~\cite{CyberEntRel,OpenCyKG} & 0.90 & 0.91  & 0.90 & \textbf{0.51
}& \textbf{0.51} & \textbf{0.49} & 0.80 & 0.80& 0.80 & 0.97 & 0.98 & 0.97 & \textbf{0.97}  & \textbf{0.98} & \textbf{0.97} \\ \hline
SecureBERT-DCNN-CRF & 0.89 & 0.91 & 0.89 & \textbf{0.41} &\textbf{0.42} &\textbf{0.42}  & 0.79 & 0.76 & 0.77 &0.97 & 0.98 & 0.97  & \textbf{0.97} & \textbf{0.98} &\textbf{0.96} \\ \hline
SecureBERT-BiGRU-CRF \cite{CTiKG} & 0.92 & 0.92 &0.91  & \textbf{0.44} & \textbf{0.42} & \textbf{0.43} & 0.79 & 0.79 & 0.79 &0.98 & 0.99 &0.98 & \textbf{0.98} & \textbf{0.98} & \textbf{0.98}\\ \hline
SecureBERT$^+$-BiGRU-CRF & 0.94 & 0.93 &0.93  & \textbf{0.41} & \textbf{0.43} & \textbf{0.44} & 0.78 & 0.79 & 0.78 &0.98 & 0.99 &0.98 & \textbf{0.99} & \textbf{0.99} & \textbf{0.98}\\ \hline

\hline
\end{tabular}
\label{tab:je_ablation_eval}
\end{table*}
In this section, we present the results of a series of ablation experiments to analyze the individual contributions of EDFs to RE performance. Specifically, we run four configurations: (i) EDF Disabled, where both the entity mask and entity type features are removed;
(ii) Entity mask only, where only the entity position feature is enabled;
(iii) Entity type only, where only the entity type feature is enabled; and
(iv) EDF enabled, where both features are active.\\
The results of these experiments are presented in Table~\ref{tab:je_ablation_eval}. Notably, the NER performance and the results under the EDF enabled and disabled settings remain consistent with those reported in Table~\ref{tab:je_eval}. However, the ablation findings in Table~\ref{tab:je_ablation_eval} provide crucial insights, including the following:

\begin{itemize}
\item \textbf{Entity Mask Only:} Enabling only the entity position feature through masking yields a moderate improvement in RE performance compared to when both features are disabled. While this helps the model localize entity boundaries, it still lacks the semantic context necessary for accurate relation classification. This is reflected in the average F1 scores, which rise modestly above 0.79.

\item \textbf{Entity Type Only:} Using only the entity type feature yields substantial performance gains, often approaching those of the full EDF enabled setup. This suggests that entity semantics are highly predictive of relation types. Given that the model predicts a relation for each entity pair in the MSLR, the entity type feature (e.g., ["HackOrg", "Tool"]) provides crucial contextual cues. As shown in Table~\ref{tab:je_ablation_eval}, average F1 scores quickly rise to ~0.97, confirming strong generalization. This setup mirrors real-world reasoning, where certain entity type pairs (e.g., "HackOrg" and "Tool") strongly imply specific relations including "uses", even without full sentence context. However, this advantage diminishes when using only raw entity texts (e.g., "APT29" and "Mimikatz"), which require domain-specific semantic understanding within the cybersecurity context.    
\end{itemize}
Overall, although the entity type feature proves to be the most influential individually, the EDF enabled configuration is particularly effective owing to its complementary design. Specifically:
(i) the entity mask enables context-aware entity representations by pooling token embeddings from the shared encoder (SecureBERT$^+$-BiGRU) for each entity pair (see Eq.~\ref{eq:pooling}), which helps disambiguate meaning in longer or syntactically complex sentences; and (ii) concatenating these contextual embeddings with the static entity type embeddings enriches the ability of the model to understand entity roles, ultimately resulting in more accurate and informed relation classification

\subsubsection{Use Cases}
Table \ref{tab:relTriples}  presents the effectiveness of TIJERE in extracting relation triples from sample sentences based on MITRE ATT\&CK~\cite{MITRE}. The model effectively and accurately extracts entities-relation triples including overlapping relations from complex sentences. Notably, entities such as "Oil", "Energy", and "Petrochemical companies" are identified separately by the model as "Org" type entities, but are grouped in one line in the table for efficient use of space. The same approach is applied to "Area" entities such as "Kazakhstan", "Taiwan", "Greece", and "the United States". The extracted entity-relation triples can be stored in a graph database such as Neo4j to construct CKGs, as shown in our previous work~\cite{TiKG,CTiKG}. These CKGs enable a comprehensive view of the security landscape and support advanced functionalities such as graph queries and link prediction that help security analysts automate threat detection and analysis, enhance situational awareness, anticipate potential attacks, and design timely; and effective defense strategies.
\begin{table*}[htp]
\centering
\caption{Sample extraction triples with TIJERE.}
\begin{tabular}{p{.3cm}|p{6.cm}|p{9.9cm}}
\hline
\multicolumn{2}{c||}{ \textbf{Sentence with highlighted threat entities}} & \textbf{Relation Triples}\\ \hline
1 & In this same time frame, \textbf{APT10} also targeted a \textbf{U.S. law firm} and an international \textbf{apparel company}, likely to \textbf{gather information} for commercial advantage.  & \makecell{(APT10, targets, U.S. law firm)\\
(APT10, targets, apparel company) \\ (APT10, motivates, gather information)}  \\ \hline
 2 & \textbf{Carbanak} is a cybercriminal group that has used \textbf{Carbanak malware} to target \textbf{financial institutions} since at least \textbf{2013}.&  \makecell{(Carbanak, uses, Carbanak malware) \\ (Carbanak, targets, financial institutions) \\ (Carbanak, hasAttackTime, 2013) \\ (Carbanak malware, targets, financial institutions) \\ (Carbanak malware, hasAttackTime, 2013) } \\ \hline
  3 & \textbf{Night Dragon} was a \textbf{cyber espionage campaign} that targeted \textbf{oil, energy, petrochemical companies}, along with individuals and executives in \textbf{Kazakhstan, Taiwan, Greece, the United States} & \makecell{(Night Dragon, uses, cyber espionage campaign) \\ (Night Dragon, targets, (oil, energy, and petrochemical companies)) \\ (Night Dragon, targets, (Kazakhstan, Taiwan, Greece, the United States)) \\ (cyber espionage campaign, targets, (oil, energy, and petrochemical \\companies)) \\ (cyber espionage campaign, targets, (Kazakhstan, Taiwan, Greece, \\the United States))}\\ \hline
\end{tabular}
\label{tab:relTriples}
\end{table*}
\subsection{Theoretical Implications and Limitations}
TIJERE introduces a novel data-driven approach that integrates EDFs, namely entity type and entity mask, to enhance entity and RE in a JE setting. These features encode essential contextual, semantic, and positional information, guiding the model to make more informed classification decisions. The MSLR serves as the mechanism that structures this information, enabling the model to effectively mitigate key JE challenges, such as overlapping relations, feature confusion, and ambiguous relation types, ultimately enhancing relation classification accuracy.\\
A key strength of this approach lies in its adaptability across various domains, including bioinformatics, finance, and healthcare. It can be applied in any field where NER or RE is relevant, as it operates independently of the tagging format used (BIO, IOB, BIOES, etc.). Entities are consistently identified by their predefined entity types or categories, which form the foundation of this innovative approach. 

The primary limitation of TIJERE is its reliance on expert knowledge to identify key features and the technical complexity involved in data transformation. However, its innovative approach offers vital insights that inform cybersecurity research and extend benefits to other fields requiring structured information extraction. Additionally, while incorporating EDF increases input dimensionality and slightly extends training time, it remains significantly faster than current state-of-the-art approaches, which often require more than half a day for training~\cite{ITIRel,CyberEntRel}. TIJERE demonstrates greater robustness, effectiveness, and efficiency, making it well-suited for constructing high-quality KGs that support real-world cybersecurity applications and enhance the workflow and decision-making of security professionals.

\section{Conclusions}  
This study introduces TIJERE, a novel framework for jointly extracting entities and their relationships from unstructured threat intelligence reports. By formulating JE as MSLR, TIJERE integrates EDFs that enrich positional, contextual, and semantic entity information, enabling the model to learn distinct features for each subtask and make more informed predictions. This data-driven approach effectively addresses key JE challenges, including overlapping relations, feature confusion, and ambiguous relation types, while enhancing the interaction between NER and RE. Experimental evaluations on the curated scalable DNRTI-JE dataset demonstrate that TIJERE outperforms state-of-the-art models, achieving an F1 score of over 0.93 for NER and an F1 score of 0.98 for RE, even with a dataset containing a high number of entities and relations. By leveraging SecureBERT$^+$, TIJERE enhances entity recognition and reduces language ambiguity in cybersecurity texts. Despite its advantages, TIJERE requires expert knowledge and data modeling to effectively integrate domain features. Future work will explore comparative evaluations between PE and JE approaches while expanding the adaptability of TIJERE to cross-domain applications such as finance, healthcare, and bioinformatics. Extending the DNRTI-JE dataset to include recent APT groups and attack scenarios is a promising research direction. This enhancement would increase dataset relevance and facilitate advanced research in AI-driven threat intelligence automation. Looking ahead, a key direction is the integration of TIJERE into real-time CTI pipelines and interactive cybersecurity dashboards, enabling security analysts to directly leverage automated extractions for proactive defense and operational decision-making.

\section*{Availability}
The DNRTI-JE dataset used for evaluation is publicly available on GitHub\footnote{\url{https://github.com/imouiche/TIJERE}}. The source code for the TIJERE framework will be made available upon reasonable request. Interested researchers may contact the corresponding author for access.



\printcredits

\bibliographystyle{cas-model2-names}

\bibliography{cas-refs}

\begin{thebibliography}{29}
\expandafter\ifx\csname natexlab\endcsname\relax\def\natexlab#1{#1}\fi
\providecommand{\url}[1]{\texttt{#1}}
\providecommand{\href}[2]{#2}
\providecommand{\path}[1]{#1}
\providecommand{\DOIprefix}{doi:}
\providecommand{\ArXivprefix}{arXiv:}
\providecommand{\URLprefix}{URL: }
\providecommand{\Pubmedprefix}{pmid:}
\providecommand{\doi}[1]{\href{http://dx.doi.org/#1}{\path{#1}}}
\providecommand{\Pubmed}[1]{\href{pmid:#1}{\path{#1}}}
\providecommand{\bibinfo}[2]{#2}
\ifx\xfnm\relax \def\xfnm[#1]{\unskip,\space#1}\fi
\bibitem[{Aghaei et~al.(2023)Aghaei, Niu, Shadid and Al-Shaer}]{SecureBERT}
\bibinfo{author}{Aghaei, E.}, \bibinfo{author}{Niu, X.},
  \bibinfo{author}{Shadid, W.}, \bibinfo{author}{Al-Shaer, E.},
  \bibinfo{year}{2023}.
\newblock \bibinfo{title}{Securebert: A domain-specific language model for
  cybersecurity}.
\newblock \bibinfo{journal}{Security and Privacy in Communication Networks}
  \bibinfo{volume}{462}.
\newblock \DOIprefix\doi{10.1109/TrustCom50675.2020.00083}.
\bibitem[{Ahmed et~al.(2024)Ahmed, Khurshid and Hina}]{CyberEntRel}
\bibinfo{author}{Ahmed, K.}, \bibinfo{author}{Khurshid, S.K.},
  \bibinfo{author}{Hina, S.}, \bibinfo{year}{2024}.
\newblock \bibinfo{title}{Cyberentrel: Joint extraction of cyber entities and
  relations using deep learning}.
\newblock \bibinfo{journal}{Computers \& Security} \bibinfo{volume}{136}.
\newblock \DOIprefix\doi{https://doi.org/10.1016/j.cose.2023.103579}.
\bibitem[{Bekoulis et~al.(2018)Bekoulis, Deleu, Demeester and
  Develder}]{GiannisBekoulis}
\bibinfo{author}{Bekoulis, G.}, \bibinfo{author}{Deleu, J.},
  \bibinfo{author}{Demeester, T.}, \bibinfo{author}{Develder, C.},
  \bibinfo{year}{2018}.
\newblock \bibinfo{title}{Joint entity recognition and relation extraction as a
  multi-head selection problem.}
\newblock \bibinfo{journal}{Expert Systems with Applications}
  \bibinfo{volume}{114}, \bibinfo{pages}{34--45}.
\bibitem[{Ehsan et~al.(2022)Ehsan, Niu, Waseem and Ehab}]{SecureBERTPlus}
\bibinfo{author}{Ehsan, A.}, \bibinfo{author}{Niu, X.},
  \bibinfo{author}{Waseem, S.}, \bibinfo{author}{Ehab, A.S.},
  \bibinfo{year}{2022}.
\newblock \bibinfo{title}{Securebert\_plus}.
\newblock \URLprefix \url{https://huggingface.co/ehsanaghaei/SecureBERT_Plus}.
\bibitem[{Gasmi et~al.(2019)Gasmi, Laval and Bouras}]{HoussemGasmi}
\bibinfo{author}{Gasmi, H.}, \bibinfo{author}{Laval, J.},
  \bibinfo{author}{Bouras, A.}, \bibinfo{year}{2019}.
\newblock \bibinfo{title}{Information extraction of cybersecurity concepts: an
  lstm approach}.
\newblock \bibinfo{journal}{Applied Sciences} \bibinfo{volume}{9},
  \bibinfo{pages}{3945}.
\newblock \DOIprefix\doi{https://doi.org/10.3390/app9193945}.
\bibitem[{Guo et~al.(2021)Guo, Liu, Huang, Liu, Jing, Wang and Wang}]{CyberRel}
\bibinfo{author}{Guo, Y.}, \bibinfo{author}{Liu, Z.}, \bibinfo{author}{Huang,
  C.}, \bibinfo{author}{Liu, J.}, \bibinfo{author}{Jing, W.},
  \bibinfo{author}{Wang, Z.}, \bibinfo{author}{Wang, Y.}, \bibinfo{year}{2021}.
\newblock \bibinfo{title}{Cyberrel: Joint entity and relation extraction for
  cybersecurity concepts}.
\newblock \bibinfo{journal}{Information and Communications Security: 23rd
  International Conference, ICICS 2021, Chongqing, China, November 19-21} ,
  \bibinfo{pages}{447–463}\DOIprefix\doi{https://doi.org/10.1007/978-3-030-86890-1_25}.
\bibitem[{Jo et~al.(2022)Jo, Lee and Shin}]{Vulcan}
\bibinfo{author}{Jo, H.}, \bibinfo{author}{Lee, Y.}, \bibinfo{author}{Shin,
  S.}, \bibinfo{year}{2022}.
\newblock \bibinfo{title}{Vulcan: Automatic extraction and analysis of cyber
  threat intelligence from unstructured text}.
\newblock \bibinfo{journal}{Computers \& Security} \bibinfo{volume}{120},
  \bibinfo{pages}{13}.
\newblock \DOIprefix\doi{10.1016/j.cose.2022.102763}.
\bibitem[{Li et~al.(2022)Li, Guo, Fang, Liu and Chen}]{YongfeiLi}
\bibinfo{author}{Li, Y.}, \bibinfo{author}{Guo, Y.}, \bibinfo{author}{Fang,
  C.}, \bibinfo{author}{Liu, Y.}, \bibinfo{author}{Chen, Q.},
  \bibinfo{year}{2022}.
\newblock \bibinfo{title}{A novel threat intelligence information extraction
  system combining multiple models}.
\newblock \bibinfo{journal}{Security and Communication Networks}
  \DOIprefix\doi{https://doi.org/10.1155/2022/8477260}.
\bibitem[{Liu et~al.(2024)Liu, Han, Zuo, Lv and Guo}]{CTI-JE}
\bibinfo{author}{Liu, Y.}, \bibinfo{author}{Han, X.}, \bibinfo{author}{Zuo,
  W.}, \bibinfo{author}{Lv, H.}, \bibinfo{author}{Guo, J.},
  \bibinfo{year}{2024}.
\newblock \bibinfo{title}{Cti-je: A joint extraction framework of entities and
  relations in unstructured cyber threat intelligence}.
\newblock \bibinfo{journal}{27th International Conference on Computer Supported
  Cooperative Work in Design (CSCWD), Tianjin, China} ,
  \bibinfo{pages}{2728--2733}\DOIprefix\doi{10.1109/CSCWD61410.2024.10580210}.
\bibitem[{Lv et~al.(2024)Lv, Han, Cui1, Wang, Zuo and Zhou}]{CTI-TFN}
\bibinfo{author}{Lv, H.}, \bibinfo{author}{Han, X.}, \bibinfo{author}{Cui1,
  H.}, \bibinfo{author}{Wang, P.}, \bibinfo{author}{Zuo, W.},
  \bibinfo{author}{Zhou, Y.}, \bibinfo{year}{2024}.
\newblock \bibinfo{title}{Joint extraction of entities and relationships from
  cyber threat intelligence based on task-specific fourier network}.
\newblock \bibinfo{journal}{2024 International Joint Conference on Neural
  Networks (IJCNN), Yokohama, Japan} ,
  \bibinfo{pages}{1--8}\DOIprefix\doi{10.1109/IJCNN60899.2024.10650942}.
\bibitem[{Marchiori et~al.(2023)Marchiori, Conti and Verde}]{STIXnet}
\bibinfo{author}{Marchiori, F.}, \bibinfo{author}{Conti, M.},
  \bibinfo{author}{Verde, N.V.}, \bibinfo{year}{2023}.
\newblock \bibinfo{title}{Stixnet: A novel and modular solution for extracting
  all stix objects in cti reports}.
\newblock \bibinfo{journal}{ARES 23: Proceedings of the 18th International
  Conference on Availability, Reliability and Security} \bibinfo{volume}{2},
  \bibinfo{pages}{1--11}.
\newblock \DOIprefix\doi{https://doi.org/10.1145/3600160.3600182}.
\bibitem[{MITRE(2023)}]{MITRE}
\bibinfo{author}{MITRE}, \bibinfo{year}{2023}.
\newblock \bibinfo{title}{Att\&ck}.
\newblock \URLprefix \url{https://attack.mitre.org/versions/v13/}.
\bibitem[{Miwa and Bansal(2016)}]{MakotoMiwa}
\bibinfo{author}{Miwa, M.}, \bibinfo{author}{Bansal, M.}, \bibinfo{year}{2016}.
\newblock \bibinfo{title}{End-to-end relation extraction using lstms on
  sequences and tree structures}.
\newblock \bibinfo{journal}{Proceedings of the 54th Annual Meeting of the
  Association for Computational Linguistics} \bibinfo{volume}{1},
  \bibinfo{pages}{1105–1116}.
\newblock \DOIprefix\doi{https://aclanthology.org/P16-1105}.
\bibitem[{Mouiche et~al.(2025)Mouiche, Merbouh and Saad}]{CTiKG}
\bibinfo{author}{Mouiche, I.}, \bibinfo{author}{Merbouh, H.},
  \bibinfo{author}{Saad, S.}, \bibinfo{year}{2025}.
\newblock \bibinfo{title}{Context-aware entity-relation extraction pipeline for
  threat intelligence knowledge graphs}.
\newblock \bibinfo{journal}{TechRxiv}
  \DOIprefix\doi{10.36227/techrxiv.173627493.39916970/v1}.
\bibitem[{Mouiche and Saad(2025)}]{TiKG}
\bibinfo{author}{Mouiche, I.}, \bibinfo{author}{Saad, S.},
  \bibinfo{year}{2025}.
\newblock \bibinfo{title}{Entity and relation extractions for threat
  intelligence knowledge graphs}.
\newblock \bibinfo{journal}{Computers \& Security} \bibinfo{volume}{148}.
\newblock \DOIprefix\doi{https://doi.org/10.1016/j.cose.2024.104120}.
\bibitem[{Piplai et~al.(2020)Piplai, Mittal, Joshi, Finin, Holt and
  Zak}]{AritranPiplai}
\bibinfo{author}{Piplai, A.}, \bibinfo{author}{Mittal, S.},
  \bibinfo{author}{Joshi, A.}, \bibinfo{author}{Finin, T.},
  \bibinfo{author}{Holt, J.}, \bibinfo{author}{Zak, R.}, \bibinfo{year}{2020}.
\newblock \bibinfo{title}{Creating cybersecurity knowledge graphs from malware
  after action reports}.
\newblock \bibinfo{journal}{in IEEE Access} ,
  \bibinfo{pages}{211691--211703}\DOIprefix\doi{10.1109/ACCESS.2020.3039234}.
\bibitem[{Sarhan and Spruit(2021)}]{OpenCyKG}
\bibinfo{author}{Sarhan, I.}, \bibinfo{author}{Spruit, M.},
  \bibinfo{year}{2021}.
\newblock \bibinfo{title}{Open-cykg: An open cyber threat intelligence
  knowledge graph}.
\newblock \bibinfo{journal}{Knowledge-Based Systems} \bibinfo{volume}{233}.
\newblock \DOIprefix\doi{https://doi.org/10.1016/j.knosys.2021.107524}.
\bibitem[{Syed et~al.(2016)Syed, Padia, Finin, Mathews and Joshi}]{UCO}
\bibinfo{author}{Syed, Z.}, \bibinfo{author}{Padia, A.},
  \bibinfo{author}{Finin, T.}, \bibinfo{author}{Mathews, L.},
  \bibinfo{author}{Joshi, A.}, \bibinfo{year}{2016}.
\newblock \bibinfo{title}{Uco: A unified cybersecurity ontology}.
\newblock \bibinfo{journal}{AAAI Workshop on Artificial Intelligence for Cyber
  Security} , \bibinfo{pages}{1--6}.
\bibitem[{Wang et~al.(2020)Wang, Liu, Ao, Li, Jiang, Xu, Xiong, Xiong and
  Zhang}]{DNRTI}
\bibinfo{author}{Wang, X.}, \bibinfo{author}{Liu, X.}, \bibinfo{author}{Ao,
  S.}, \bibinfo{author}{Li, N.}, \bibinfo{author}{Jiang, Z.},
  \bibinfo{author}{Xu, Z.}, \bibinfo{author}{Xiong, Z.},
  \bibinfo{author}{Xiong, M.}, \bibinfo{author}{Zhang, X.},
  \bibinfo{year}{2020}.
\newblock \bibinfo{title}{Dnrti: A large-scale dataset for named entity
  recognition in threat intelligence}.
\newblock \bibinfo{journal}{2020 IEEE 19th International Conference on Trust,
  Security, and Privacy in Computing and Communications (TrustCom), Guangzhou,
  China} ,
  \bibinfo{pages}{1842--1848}\DOIprefix\doi{https://doi.org/10.1109/TrustCom50675.2020.00252}.
\bibitem[{Wang et~al.(2024)Wang, Liu and Liu}]{XiaodiWang}
\bibinfo{author}{Wang, X.}, \bibinfo{author}{Liu, Z.}, \bibinfo{author}{Liu,
  J.}, \bibinfo{year}{2024}.
\newblock \bibinfo{title}{Joint relational triple extraction with enhanced
  representation and binary tagging framework in cybersecurity}.
\newblock \bibinfo{journal}{Computers \& Security} \bibinfo{volume}{144}.
\bibitem[{Wang et~al.(2022)Wang, Zhang and Deng}]{BERT-CRF}
\bibinfo{author}{Wang, Y.}, \bibinfo{author}{Zhang, X.}, \bibinfo{author}{Deng,
  A.}, \bibinfo{year}{2022}.
\newblock \bibinfo{title}{Joint extraction model of entity relations based on
  bert-crf}.
\newblock \bibinfo{journal}{2022 4th International Conference on Machine
  Learning, Big Data and Business Intelligence (MLBDBI), Shanghai, China} ,
  \bibinfo{pages}{7--12}\DOIprefix\doi{10.1109/MLBDBI58171.2022.00008}.
\bibitem[{Yan et~al.(2022)Yan, Jia and Tu}]{PipeVsJoint}
\bibinfo{author}{Yan, Z.}, \bibinfo{author}{Jia, Z.}, \bibinfo{author}{Tu, K.},
  \bibinfo{year}{2022}.
\newblock \bibinfo{title}{An empirical study of pipeline vs. joint approaches
  to entity and relation extraction}.
\newblock \bibinfo{journal}{In Proceedings of the 2nd Conference of the
  Asia-Pacific Chapter of the Association for Computational Linguistics and the
  12th International Joint Conference on Natural Language Processing} ,
  \bibinfo{pages}{437–443}.
\bibitem[{You et~al.(2024)You, Jiang, Zhang, Feng, Jiang and Yang}]{TiGNet}
\bibinfo{author}{You, Y.}, \bibinfo{author}{Jiang, Z.}, \bibinfo{author}{Zhang,
  K.}, \bibinfo{author}{Feng, H.}, \bibinfo{author}{Jiang, J.},
  \bibinfo{author}{Yang, P.}, \bibinfo{year}{2024}.
\newblock \bibinfo{title}{Tignet: Joint entity and relation triplets extraction
  for apt campaign threat intelligence}.
\newblock \bibinfo{journal}{27th International Conference on Computer Supported
  Cooperative Work in Design (CSCWD), Tianjin, China} ,
  \bibinfo{pages}{1687--1694}\DOIprefix\doi{10.1109/CSCWD61410.2024.10580395}.
\bibitem[{Yuan et~al.(2020)Yuan, Zhou, Pan, Zhu, Song and Guo}]{YueYuan}
\bibinfo{author}{Yuan, Y.}, \bibinfo{author}{Zhou, X.}, \bibinfo{author}{Pan,
  S.}, \bibinfo{author}{Zhu, Q.}, \bibinfo{author}{Song, Z.},
  \bibinfo{author}{Guo, L.}, \bibinfo{year}{2020}.
\newblock \bibinfo{title}{A relation-specific attention network for joint
  entity and relation extraction}.
\newblock \bibinfo{journal}{In: International Joint Conference on Artificial
  Intelligence 2020. Association for the Advancement of Artificial Intelligence
  (AAAI)} , \bibinfo{pages}{4054–4060}.
\bibitem[{Zhang et~al.(2024)Zhang, Xu, Chen, Zhou and Cheng}]{HuikangZhang}
\bibinfo{author}{Zhang, H.}, \bibinfo{author}{Xu, Y.}, \bibinfo{author}{Chen,
  J.}, \bibinfo{author}{Zhou, W.}, \bibinfo{author}{Cheng, L.},
  \bibinfo{year}{2024}.
\newblock \bibinfo{title}{A knowledge graph for network security}.
\newblock \bibinfo{journal}{In Wang, W., Liu, X., Na, Z., Zhang, B. (eds)
  Communications, Signal Processing, and Systems. CSPS 2023. Lecture Notes in
  Electrical Engineering} \bibinfo{volume}{1032}.
\newblock \DOIprefix\doi{10.1007/978-981-99-7505-1\_59}.
\bibitem[{Zhao et~al.(2020)Zhao, Yan, Liu1, Li and Zuo}]{JunZhao}
\bibinfo{author}{Zhao, J.}, \bibinfo{author}{Yan, Q.}, \bibinfo{author}{Liu1,
  X.}, \bibinfo{author}{Li, B.}, \bibinfo{author}{Zuo, G.},
  \bibinfo{year}{2020}.
\newblock \bibinfo{title}{Cyber threat intelligence modeling based on
  heterogeneous graph convolutional network}.
\newblock \bibinfo{journal}{In Proceedings of the 23rd international symposium
  on research in attacks, intrusions and defenses (RAID 2020)}
  \bibinfo{volume}{1}, \bibinfo{pages}{241–256}.
\bibitem[{Zhong and Chen(2021)}]{ZexuanZhong}
\bibinfo{author}{Zhong, Z.}, \bibinfo{author}{Chen, D.}, \bibinfo{year}{2021}.
\newblock \bibinfo{title}{A frustratingly easy approach for entity and relation
  extraction}.
\newblock \bibinfo{journal}{In Proceedings of the 2021 Conference of the North
  American Chapter of the Association for Computational Linguistics: Human
  Language Technologies} , \bibinfo{pages}{50–61}.
\bibitem[{Zhu et~al.(2024)Zhu, Cheng, Li and Xu}]{ITIRel}
\bibinfo{author}{Zhu, F.}, \bibinfo{author}{Cheng, Z.}, \bibinfo{author}{Li,
  P.}, \bibinfo{author}{Xu, H.}, \bibinfo{year}{2024}.
\newblock \bibinfo{title}{Itirel: Joint entity and relation extraction for
  internet of things threat intelligence}.
\newblock \bibinfo{journal}{in IEEE Internet of Things Journal} ,
  \bibinfo{pages}{20867--20878}\DOIprefix\doi{10.1109/JIOT.2024.3373799}.
\bibitem[{Zuo et~al.(2022)Zuo, Gao, Li and Yuan}]{ZuoYali}
\bibinfo{author}{Zuo, J.}, \bibinfo{author}{Gao, Y.}, \bibinfo{author}{Li, X.},
  \bibinfo{author}{Yuan, J.}, \bibinfo{year}{2022}.
\newblock \bibinfo{title}{An end-to-end entity and relation joint extraction
  model for cyber threat intelligence}.
\newblock \bibinfo{journal}{2022 the 7th International Conference on Big Data
  Analytics (ICBDA)} \DOIprefix\doi{10.1109/ICBDA55095.2022.9760342}.

\end{thebibliography}





\end{document}